\documentclass{article}
\usepackage{spconf,amsmath,graphicx}
\usepackage{multirow}
\usepackage{booktabs}
\usepackage{graphics}
\usepackage{adjustbox}
\usepackage{graphicx}
\usepackage{subcaption}
\usepackage[font=small,labelfont=bf]{caption}

\title{A NOVEL FRAMEWORK FOR ASSESSMENT OF LEARNING-BASED DETECTORS IN REALISTIC CONDITIONS WITH APPLICATION TO DEEPFAKE DETECTION}
\name{Yuhang Lu, Ruizhi Luo, Touradj Ebrahimi\thanks{Support from XAIface
CHIST-ERA-19-XAI-011 and the Swiss National Science
Foundation (SNSF) 20CH21\_195532 is acknowledged.}}
\address{\'Ecole Polytechnique F\'ed\'erale de Lausanne (EPFL)\\
Multimedia Signal Processing Group (MMSPG)}
\begin{document}
%
\maketitle
\begin{abstract}
Deep convolutional neural networks have shown remarkable results on multiple detection tasks. Despite the significant progress, the performance of such detectors are often assessed in public benchmarks under non-realistic conditions. Specifically, impact of conventional distortions and processing operations such as compression, noise, and enhancement are not sufficiently studied. This paper proposes a rigorous framework to assess performance of learning-based detectors in more realistic situations. An illustrative example is shown under deepfake detection context. Inspired by the assessment results, a data augmentation strategy based on natural image degradation process is designed, which significantly improves the generalization ability of two deepfake detectors.
\end{abstract}
\begin{keywords}
Assessment framework, Deepfake detection, Data augmentation
\end{keywords}

\section{Introduction}

The performance of the deep convolutional neural networks (DCNN) based detectors are often measured through publicly available benchmarks. However, it has been shown that many DCNN-based methods are vulnerable to real-world perturbation and post-processing operations \cite{Dodge2016UnderstandingHI, Zhou2017OnCO, hendrycks2019robustness}.
In realistic situations, images can face unpredictable distortions from the extrinsic environment, such as noise and poor illumination conditions, or constantly undergo various processing operations to ease transmission. This could create critical challenges for safety and security applications, for instance, in surveillance, in autonomous vehicles, and in digital forensics. 

Take deepfake detection as an example. Deepfakes refer to manipulated face contents using deep learning techniques. The advanced deepfake creation algorithms and open source software make it accessible and cheap to the public, posing risks to the security and authenticity of media. 
Although current deepfake detectors have been pushing the limits of all available benchmarks, they are for most parts, developed under ideal conditions. Malicious agents could easily fool the detector by adding noise imperceptible to human eyes or by applying more aggressive compression ratios to the targeted content. 
Therefore, it is desired to design a more comprehensive and systematic approach with an in-depth analysis of how the different processing operations impact the detection performance. 

In this work, an assessment framework is presented for generic learning-based detection systems, which extensively evaluate the performance of a detector towards different real-world modifications. To generate a broad range of variations on test data in a realistic manner, common post-processing operations such as image transcoding, smoothing, enhancement, resizing, and synthetic noises are employed. The effectiveness of our framework is illustrated under the deepfake detection context, where thorough experiments are conducted and insightful conclusions are drawn. In the end, we have designed a data augmentation strategy based on realistic image degradation modeling process, which significantly improves the generalization ability of a deepfake detector at marginal impact on the performance in ideal conditions.


\section{Related Work}

Several studies have investigated the vulnerability of CNN-based model to real-world and common image corruptions. Dodge and Karam \cite{Dodge2016UnderstandingHI} first measured the performance of image classification models on data suffered from noise, contrast variation and image blur. Hendrycks et al. \cite{hendrycks2019robustness} proposed a benchmark to evaluate the robustness of image recognition models towards common corruptions. Extensive work \cite{michaelis2019dragon, 2020} has been carried out in object detection and semantic segmentation and applied to safety-critical applications. Current activities in this area mostly focus on corruptions during data acquisition and apply to only one type of computer vision task. Our proposed assessment framework offers a more general solution and in addition considers the impact of realistic image processing operations in the end-to-end workflow.   

Face manipulation detection is a classical problem in computer vision. Currently, the large majority of proposed detectors treat it as a binary classification task. Similar to other computer vision tasks, a number of large-scale datasets, benchmarks, and competitions \cite{roessler2019faceforensicspp, Celeb_DF_cvpr20, jiang2020deeperforensics10, DFDC2020} are released and organized to assist the community to resolve this problem. For instance, FaceForensics++ \cite{roessler2019faceforensicspp} is one of the most popular face forensics dataset. Facebook released one of the biggest deepfake datasets, DFDC \cite{DFDC2020}, and organized a competition based on the latter. As a results, a large amount of deepfake detection methods \cite{zhou_two-stream_2017, afchar_mesonet_2018, nguyen_use_2019, Li2020FaceXF} have been proposed. One of the first deep learning-based methods for deepfake detection was proposed by Zhou et al. in \cite{zhou_two-stream_2017}. R\"ossler et al. \cite{roessler2019faceforensicspp} proposed to finetune Xception network on face manipulation dataset and showed outstanding results on the FFpp benchmark. Nguyen et al. \cite{nguyen_use_2019} leveraged capsule network to detect face manipulation. Besides focusing on spatial domain, recent works \cite{liu2021spatialphase} attempt to resolve the problem in the frequency domain.  

\section{Proposed Assessment Framework}

In this section, we describe a rigorous assessment framework for generic detection and recognition tasks to assess their performance in more realistic conditions. It is hoped that this framework will serve as a broad benchmarking approach to evaluate robustness in realistic image processing and corruptions mechanisms while at the same time providing insights on improving the approach under assessment. In general, our framework contains six categories of processing operations or corruptions with more than ten minor types. Each type consists of over five different severity levels. The details of all operations used in evaluations are described below with the illustration of a typical example in Fig. \ref{fig:op-example}.

\textit{Compression}:  JPEG compression is included in the proposed framework with multiple compression factors. As deep learning-based compression technique are becoming increasingly popular, the technique developed by Ball\'e et al. \cite{Ball2018VariationalIC} is also included in this framework.

\textit{Smoothing}: Image blurring, also known as smoothing, is a widely employed operation to reduce noise which simultaneously results in a reduction of details. Three frequently used filters with various kernel sizes are considered in our framework, including Gaussian, Median, and Average filters.

\textit{Noise}: The acquisition of images can be easily affected by noise. Our framework applies Additive White Gaussian noise (AWGN) with 5 levels of variance. To better reflect the realistic situations, a synthetic Poissonian-Gaussian noise is also considered, the parameters of which are learned from real-world noisy images.  

\textit{Enhancement}: Image enhancement is generally a very frequently used technique of adjusting images for better display or further image analysis. We change the contrast and brightness of images by separately applying linear adjustment and Gamma correction.

\textit{Resizing}: Low-resolution data can significantly reduce the performance of modern deep learning-based detectors \cite{article, Li2019OnLF}. This is often the case when the detector is employed in an outdoor environment, where captured data could have limited resolution. In this framework, we mainly measure the impact of image downscaling.

\textit{Combinations}: A mixture of two or three operations above is also considered, such as combining JPEG compression and Gaussian noise, making the test data better reflect more complex real-world scenarios.

To validate the assessment framework, one detector should be trained on its original target datasets, such as FFpp for deepfake detection tasks. Processing operations and corruptions are not applied on training data. Furthermore, several parameters were used for processing operations to better understand their impact on the detection.

\newcommand\w{0.25\linewidth}
\newcommand\y{\linewidth}
\begin{figure}[h]
\centering
\begin{subfigure}[b]{\w}
  \includegraphics[width=\y]{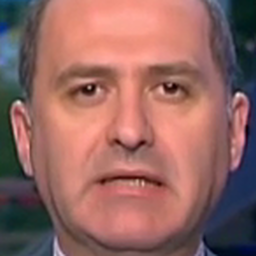}  
  \caption{Unaltered}
\end{subfigure}%
\hfill
\begin{subfigure}[b]{\w}
  \includegraphics[width=\y]{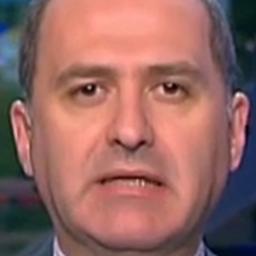}  
  \caption{JPEG}
\end{subfigure}%
\hfill
\begin{subfigure}[b]{\w}
  \includegraphics[width=\y]{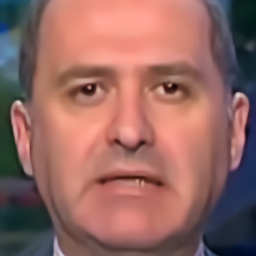}  
  \caption{DL-Comp}
\end{subfigure}%
\hfill
\begin{subfigure}[b]{\w}
  \includegraphics[width=\y]{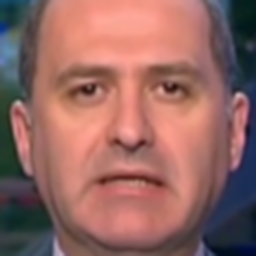}  
  \caption{GB}
\end{subfigure}%
\hfill
\begin{subfigure}[b]{\w}
  \includegraphics[width=\y]{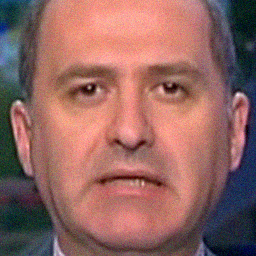}  
  \caption{GN}
\end{subfigure}%
\hfill
\begin{subfigure}[b]{\w}
  \includegraphics[width=\y]{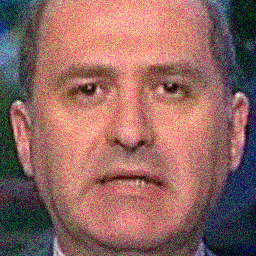}  
  \caption{Po-Gau-N}
\end{subfigure}%
\hfill
\begin{subfigure}[b]{\w}
  \includegraphics[width=\y]{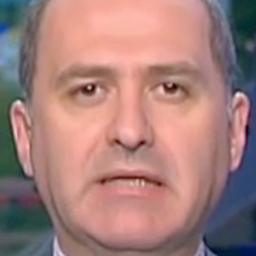}  
  \caption{Gamma}
\end{subfigure}%
\hfill
\begin{subfigure}[b]{\w}
  \includegraphics[width=\y]{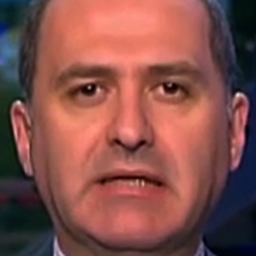}
  \caption{Gamma}
\end{subfigure}%
\hfill
\begin{subfigure}[b]{\w}
  \includegraphics[width=\y]{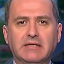}  
  \caption{Low Res}
\end{subfigure}%
\hfill
\begin{subfigure}[b]{\w}
  \includegraphics[width=\y]{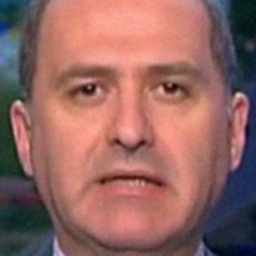}  
  \caption{GN+GB}
\end{subfigure}%
\hfill
\begin{subfigure}[b]{\w}
  \includegraphics[width=\y]{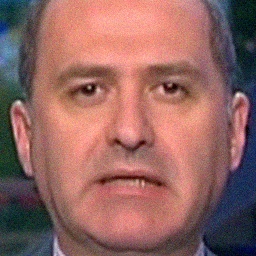}  
  \caption{JPEG+GN}
\end{subfigure}%
\hfill
\begin{subfigure}[b]{\w}
  \includegraphics[width=\y]{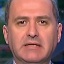}  
  \caption{JPEG+LR}
\end{subfigure}
\caption{Example of a typical frame in the test dataset after applying various operations. Some notations are explained as following. DL-Comp: learning-based compression. GB: Gaussian blur. GN: Gaussian noise. Po-Gau-N: Poissonian Gaussian noise. Gamma: Gamma correction. $+$: mixture.}
\label{fig:op-example}
\end{figure}



\section{Illustrative Example for Deepfake Detection}
The proposed framework is assessed in a deepfake detection scenario, where the robustness of two deepfake detectors are avaluated in presence of realistic processing operations. 



\subsection{Assessment Datasets}
To show the effectiveness of our assessment framework, it is applied to two widely used Deepfake detection databases.
\textbf{FaceForensics++} \cite{roessler2019faceforensicspp} is used as the main database for our assessment framework. It contains 1000 pristine and 4000 manipulated contents generated by four different approaches. Video contents are also compressed with two quality settings using the H.264, denoted as C23 and C40. Consequently, video contents corresponding to three quality levels are used in the training, while only the highest quality contents are used in the final assessment. 
\textbf{Celeb-DF} \cite{Celeb_DF_cvpr20} is a high quality face forensics dataset. Additional experiments were conducted on the latter to validate the effectiveness of our framework. The test data was selected as recommended in~\cite{Celeb_DF_cvpr20} while the validation and training data was split in 20\% versus 80\% respectively.  

\subsection{Detection Methods}
In this section, two deep learning-based deepfake detection approaches are evaluated to demonstrate the usefulness of the proposed assessment framework. Both have been reported to have outstanding performance on public benchmarks. 

\textbf{Capsule-Forensics} \cite{nguyen_use_2019}  combines traditional CNN and Capsule networks, which requires fewer parameters and at the same time achieves high detection accuracy. The model is trained from scratch on the above-mentioned two datasets. 

\textbf{XceptionNet} was first introduced in~\cite{Chollet2017XceptionDL} and became a popular CNN architecture used in multiple computer vision tasks. 
R\"ossler et al. \cite{roessler2019faceforensicspp} further adopted it for deepfake detection. It has been shown to outperform several alternative detectors in FFpp benchmark on both uncompressed and compressed contents. We leverage the model pretrained on ImageNet and finetune on the above-mentioned two datasets respectively. 


\subsection{Experiments and Results}
\subsubsection{Implementation Details}
For training, the Capsule-Forensics model was trained using the Adam optimizer with $\beta_1 = 0.9$, $\beta_2 = 0.999$, and a learning rate of $5 \times 10^{-4}$ for 25 epochs. The XceptionNet was finetuned for 10 epochs with learning rate of $1 \times 10^{-3}$. 

For both datasets, 100 frames were randomly extracted from each video for training purpose and 32 frames were extracted for validation and testing. Extracted frames were pre-processed and cropped around face regions. Processing operations and corruptions from our assessment framework were used only to test data for further evaluation. 

During the evaluation, the Accuracy (ACC), the Area Under Receiver Operating Characteristic Curve (AUC), and F1-score were used as  performance metrics in all  experiments. 


\begin{table*}[t]
  \centering
  \caption{AUC (\%) scores of two detectors tested on unaltered and distorted variants of FFpp and Celeb-DF test set respectively. Capsule-Forensics detector is shortened as \textit{Capsule}. The suffixes \textit{Raw}, \textit{C23} and \textit{Full} refer to different quality settings of FFpp.}
    \begin{adjustbox}{width=\textwidth}

    \begin{tabular}{cccccccccccccccccccccccc}
    \toprule
    \multicolumn{1}{c}{\multirow{2}[4]{*}{Methods}} & \multirow{2}[4]{*}{TrainSet} & \multirow{2}[4]{*}{} & \multirow{2}[4]{*}{Unaltered} & \multicolumn{3}{c}{JPEG} & \multicolumn{3}{c}{DL-Comp} & \multicolumn{3}{c}{Gau Noise} & \multirow{2}[4]{*}{\shortstack{Pois-Gau \\ Noise}} & \multicolumn{3}{c}{Gau Blur} & \multicolumn{4}{c}{Gamma Corr} & \multicolumn{3}{c}{Resize} \\
\cmidrule{5-13}\cmidrule{15-24}          &       &       &       & 95    & 60    & 30    & High  & Med   & Low   & 5     & 30    & 50    &       & 3     & 7     & 11    & 0.1   & 0.75  & 1.3   & 2.5   & x4    & x8    & x16 \\
\cmidrule{1-2}\cmidrule{4-24}    \multicolumn{1}{c}{\multirow{4}[1]{*}{Capsule}} & FFpp-Raw &       & 99.20 & 97.91 & 76.48 & 59.60 & 44.76 & 45.50 & 49.08 & 61.80 & 51.26 & 49.16 & 55.63 & 67.19 & 41.78 & 47.74 & 49.50 & 98.86 & 99.17 & 96.12 & 55.42 & 47.82 & 46.90 \\
          & FFpp-C23 &       & 96.32 & 95.09 & 95.76 & 74.91 & 43.04 & 42.58 & 81.57 & 84.51 & 58.63 & 50.51 & 70.59 & 85.21 & 53.94 & 47.96 & 47.92 & 95.06 & 96.72 & 92.91 & 79.33 & 64.62 & 50.33 \\
          & FFpp-Full &       & 94.52 & 94.95 & 93.97 & 84.50 & 99.01 & 96.77 & 88.95 & 89.03 & 57.95 & 51.11 & 64.87 & 85.72 & 58.83 & 56.05 & 56.02 & 93.86 & 93.87 & 85.44 & 87.05 & 69.93 & 54.15 \\
          & Celeb-DF &       & 99.76 & 99.80 & 99.33 & 96.51 & 99.01 & 96.77 & 88.95 & 97.35 & 63.30 & 44.67 &   -    & 99.15 & 96.54 & 90.58 & 48.33 & 99.71 & 99.71 & 93.44 & 95.67 & 75.16 & 68.35 \\
    \multicolumn{1}{r}{\multirow{2}[1]{*}{XceptionNet}} & FFpp-Raw &       & 99.56 & 76.77 & 56.00 & 54.20 & 50.16 & 50.37 & 50.10 & 50.12 & 49.64 & 49.30 & 48.98 & 68.76 & 55.61 & 50.70 & 54.66 & 98.66 & 99.57 & 70.45 & 68.60 & 55.80 & 50.45 \\
          & Celeb-DF &       & 98.06 & 98.20  & 97.63 & 94.98 & 96.23   & 90.23  & 75.46   & 95.92 & 63.19 & 55.93 &  \multicolumn{1}{c}{-} & 97.32 & 87.22 & 78.05 & 53.25 & 97.63 & 98.34 & 89.02 & 85.47 & 59.40  & 49.21 \\
\cmidrule{1-2}\cmidrule{4-24}    Capsule & FFpp-Raw+Aug &       & 98.16 & 97.97 & 96.36 & 94.08 & 93.81 & 71.41 & 59.74 & 97.05 & 83.51 & 75.09 & 90.04 & 96.86 & 90.32 & 80.31 & 60.17 & 97.68 & 98.18 & 96.91 & 93.54 & 79.22 & 58.05 \\
    XceptionNet & FFpp-Raw+Aug &       & 98.44 & 98.25 & 97.36 & 96.12 & 98.03 & 87.76 & 82.74 & 97.37 & 91.71 & 88.70 & 94.57 & 98.31 & 97.35 & 94.51 & 80.48 & 98.25 & 98.44 & 97.75 & 97.30 & 86.26 & 67.14 \\
    \bottomrule
    \end{tabular}%
    \end{adjustbox}
  \label{tab:result}%
\end{table*}%

\begin{figure}[h]
	\centering
    \includegraphics[width=0.9\linewidth]{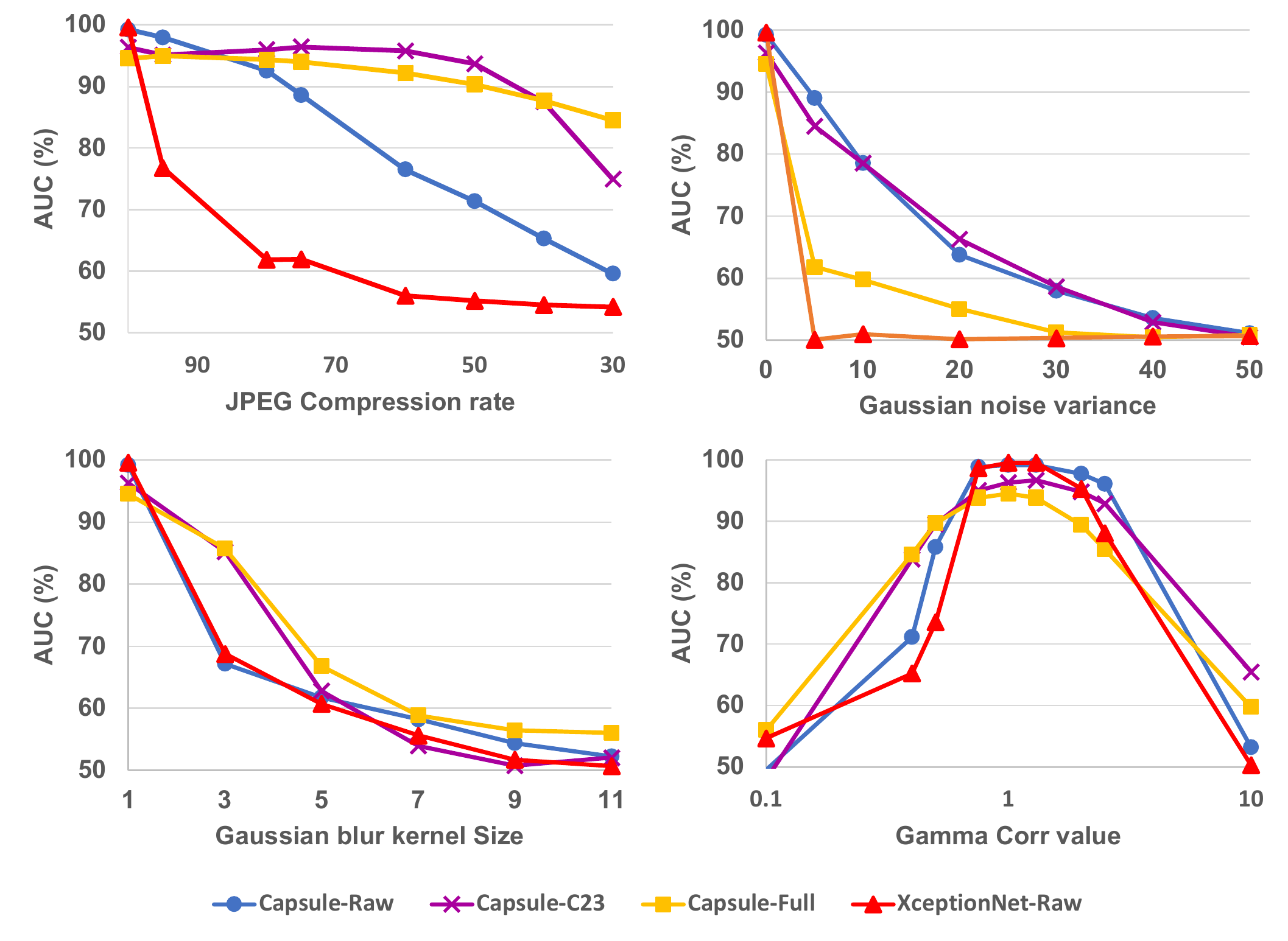}
	\caption{Assessment results of two models trained on FFpp dataset. The suffixes of legends refer to the qualities of the training data. \textit{Full} means using all available quality, for training}
	\label{fig:af}
\end{figure}


\subsubsection{Assessment Results}

The two deepfake detectors were trained on the original unaltered training sets of both FFpp and Celeb-DF. Table \ref{tab:result} shows the evaluation results using our assessment framework. Due to the page limit, only AUC scores and a subset of operations and intensity values are presented in this paper. 

In general, our findings show that even mild real-world processing operations or perturbations can have an obvious negative impact on the detection accuracy. For example, the two learning-based detectors present exceptional performance on unaltered FFpp testing data as expected, but then show severe performance degradation on modified data from our assessment framework, which indicates a lack of robustness. There is a similar pattern from the experiment results on Celeb-DF dataset, demonstrating the effectiveness of our assessment framework.

The assessment results are visualized in Figure \ref{fig:af}. One can observe that natural noise and Gaussian blur are very prominent influencing factors for deepfake detectors. The performance of both models deteriorates rapidly after increasing noise variance or filter kernel size. Capsule-Forensics detector remains relatively stable while facing mild JPEG compression and gamma correction operation, but sill breaks down as the processing intensity grows. 

In addition, the impact of quality variants of training data on learning-based detector has been analyzed. The Capsule-Forensics model trained only with very high quality data (FFpp-Raw) will be extremely sensitive to nearly all kinds of realistic processing operations. On the contrary, although mixing relatively lower quality data during the training process will decrease the accuracy on unaltered test set, it improves the robustness towards low-intensity processing and distortions, particularly for JPEG compression.

\begin{figure}[h]
	\centering
    \includegraphics[width=0.9\linewidth]{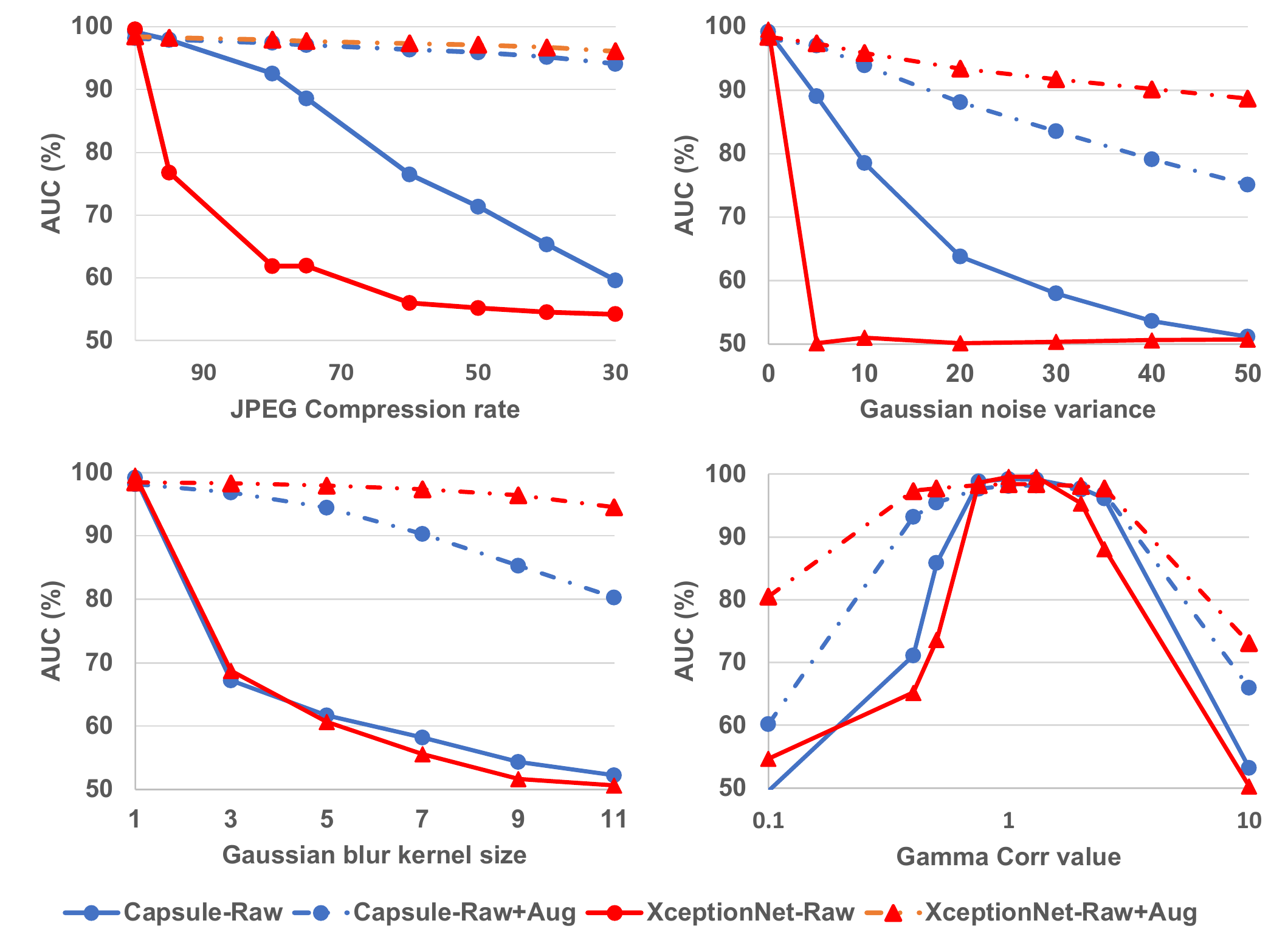}
	\caption{Performance comparison between model trained on FFpp-Raw only and trained with the proposed augmentation scheme.}
	\label{fig:aug}
\end{figure}


\section{Improved Training Strategy by Data Augmentation}


To reduce the strong negative impact of real-world processing and corruptions on model performance observed above, we propose a simple yet efficient data augmentation approach that leads to a robustness improvement.

Previous experimental results suggest that properly mixing low quality training data, slightly improves the generalization ability of detectors. Some researchers have explored using Gaussian noise corrupted data \cite{Ford2019AdversarialEA, Rusak2020IncreasingTR} or stylized data \cite{michaelis2019dragon} for training to improve model performance under corruption. However, according to our experiments with deepfake detectors, Gaussian noise-based augmentation only benefits limited type of realistic corruption and meanwhile deteriorates the performance on original unaltered data.
Based on the observation of data distortions in realistic conditions, a carefully designed augmentation chain was conceived. 
It mainly includes 4 realistic data degradation types with various intensity levels. Each process will be randomly applied to certain amount of training data in a natural sequence as following: 

\textit{Enhancement}: At the beginning of the augmentation chain, there is 50\% probability that either the brightness or the contrast of the training data will be non-linearly modified by a factor randomly sampled from $[0.5, 1.5]$. 

\textit{Smoothing}: For each batch of training data, an image blurring technique will be applied with probability of 50\%. Either Gaussian blur or Average blur filter is used with a kernel size varying from $[3, 15]$. 

\textit{Additive Gaussian Noise}: Gaussian noise is added in the augmentation chain with probability of 30\%. The standard deviation value is randomly sampled from $[0, 50]$.

\textit{JPEG Compression }: In the end, JPEG compression is applied with a probability of 70\%. Each time the quality factor is randomly sampled from $[10, 95]$.



\begin{table}[htbp]
  \centering
  \caption{Cross-dataset evaluation on Celeb-DF (AUC(\%)) after training on FFpp. The suffixes denote training set quality.}
    \begin{tabular}{cccc}
    \toprule
    Methods &       & \multicolumn{1}{c}{FFpp} & \multicolumn{1}{c}{Celeb-DF} \\
\cmidrule{1-1}\cmidrule{3-4}    Capsule-Raw &       & 99.20 & 46.70 \\
    Capsule-C23 &       & 96.32 & 52.61 \\
    Capsule-Full &       & 94.52 & 67.88 \\
    Capsule-Raw+Aug &       & 98.16 & \textbf{74.84} \\
\cmidrule{1-1}\cmidrule{3-4}    Xception-Raw &       & 99.56 & 50.70 \\
    Xception-C23 &       & 98.71      & 55.74 \\
    Xception-Raw+Aug &       & 98.44 & \textbf{80.67} \\
    \bottomrule
    \end{tabular}%
  \label{tab:cross}%
\end{table}%

The overall results are presented in the bottom of Table \ref{tab:result} which is denoted by suffix \textit{+Aug}. It is noticeable that training with our augmentation technique remarkably improves the performance on nearly all kinds of processed data even with extreme intensity values. Figure \ref{fig:aug} illustrates the performance improvement of two models on four types of distortions and processing after using our augmentation scheme.



Moreover, we evaluate the generalization ability of the models trained with our new augmentation scheme on unseen data via cross-dataset assessment. The results are shown in Table \ref{tab:cross}. The models are trained on FFpp dataset but tested on Celeb-DF. Our augmentation scheme brings significant performance improvement on both detectors while maintaining high accuracy on original dataset.

\section{Conclusion}

Many detectors are designed to be an as high performing as possible on specific benchmarks. But this often results in sacrificing generalization ability to more realistic situations. The proposed assessment framework is capable of assessing detectors in more realistic conditions and provides valuable insights on designing more robust techniques. A carefully conceived augmentation chain based on a natural data degradation process is proposed and significantly improves the model robustness against various distortions.

\bibliographystyle{IEEEbib}
\bibliography{refs}

\begin{thebibliography}{10}

\bibitem{Dodge2016UnderstandingHI}
Samuel~F. Dodge and Lina Karam,
\newblock ``Understanding how image quality affects deep neural networks,''
\newblock {\em 2016 Eighth International Conference on Quality of Multimedia
  Experience (QoMEX)}, pp. 1--6, 2016.

\bibitem{Zhou2017OnCO}
Yiren Zhou, Sibo Song, and Ngai-Man Cheung,
\newblock ``On classification of distorted images with deep convolutional
  neural networks,''
\newblock {\em 2017 IEEE International Conference on Acoustics, Speech and
  Signal Processing (ICASSP)}, pp. 1213--1217, 2017.

\bibitem{hendrycks2019robustness}
Dan Hendrycks and Thomas Dietterich,
\newblock ``Benchmarking neural network robustness to common corruptions and
  perturbations,''
\newblock {\em Proceedings of the International Conference on Learning
  Representations}, 2019.

\bibitem{michaelis2019dragon}
Claudio Michaelis, Benjamin Mitzkus, Robert Geirhos, Evgenia Rusak, Oliver
  Bringmann, Alexander~S. Ecker, Matthias Bethge, and Wieland Brendel,
\newblock ``Benchmarking robustness in object detection: Autonomous driving
  when winter is coming,''
\newblock {\em arXiv preprint arXiv:1907.07484}, 2019.

\bibitem{2020}
Christoph Kamann and Carsten Rother,
\newblock ``Benchmarking the robustness of semantic segmentation models,''
\newblock {\em 2020 IEEE/CVF Conference on Computer Vision and Pattern
  Recognition (CVPR)}, Jun 2020.

\bibitem{roessler2019faceforensicspp}
Andreas R\"ossler, Davide Cozzolino, Luisa Verdoliva, Christian Riess, Justus
  Thies, and Matthias Nie{\ss}ner,
\newblock ``Face{F}orensics++: Learning to detect manipulated facial images,''
\newblock in {\em International Conference on Computer Vision (ICCV)}, 2019.

\bibitem{Celeb_DF_cvpr20}
Yuezun Li, Xin Yang, Pu~Sun, Honggang Qi, and Siwei Lyu,
\newblock ``Celeb-df: A large-scale challenging dataset for deepfake
  forensics,''
\newblock in {\em IEEE Conference on Computer Vision and Patten Recognition
  (CVPR)}, 2020.

\bibitem{jiang2020deeperforensics10}
Liming Jiang, Ren Li, Wayne Wu, Chen Qian, and Chen~Change Loy,
\newblock ``Deeperforensics-1.0: A large-scale dataset for real-world face
  forgery detection,'' 2020.

\bibitem{DFDC2020}
Brian Dolhansky, Joanna Bitton, Ben Pflaum, Jikuo Lu, Russ Howes, Menglin Wang,
  and Cristian~Canton Ferrer,
\newblock ``The deepfake detection challenge dataset,'' 2020.

\bibitem{zhou_two-stream_2017}
Peng Zhou, Xintong Han, Vlad~I. Morariu, and Larry~S. Davis,
\newblock ``Two-{Stream} {Neural} {Networks} for {Tampered} {Face}
  {Detection},''
\newblock in {\em 2017 {IEEE} {Conference} on {Computer} {Vision} and {Pattern}
  {Recognition} {Workshops} ({CVPRW})}, July 2017, pp. 1831--1839,
\newblock ISSN: 2160-7516.

\bibitem{afchar_mesonet_2018}
Darius Afchar, Vincent Nozick, Junichi Yamagishi, and Isao Echizen,
\newblock ``{MesoNet}: a {Compact} {Facial} {Video} {Forgery} {Detection}
  {Network},''
\newblock in {\em 2018 {IEEE} {International} {Workshop} on {Information}
  {Forensics} and {Security} ({WIFS})}, Dec. 2018, pp. 1--7,
\newblock ISSN: 2157-4774.

\bibitem{nguyen_use_2019}
Huy~H. Nguyen, Junichi Yamagishi, and Isao Echizen,
\newblock ``Use of a {Capsule} {Network} to {Detect} {Fake} {Images} and
  {Videos},''
\newblock {\em arXiv:1910.12467 [cs]}, Oct. 2019,
\newblock arXiv: 1910.12467.

\bibitem{Li2020FaceXF}
Lingzhi Li, Jianmin Bao, Ting Zhang, Hao Yang, Dong Chen, Fang Wen, and Baining
  Guo,
\newblock ``Face x-ray for more general face forgery detection,''
\newblock {\em 2020 IEEE/CVF Conference on Computer Vision and Pattern
  Recognition (CVPR)}, pp. 5000--5009, 2020.

\bibitem{liu2021spatialphase}
Honggu Liu, Xiaodan Li, Wenbo Zhou, Yuefeng Chen, Yuan He, Hui Xue, Weiming
  Zhang, and Nenghai Yu,
\newblock ``Spatial-phase shallow learning: Rethinking face forgery detection
  in frequency domain,'' 2021.

\bibitem{Ball2018VariationalIC}
Johannes Ball{\'e}, David~C. Minnen, Saurabh Singh, Sung~Jin Hwang, and Nick
  Johnston,
\newblock ``Variational image compression with a scale hyperprior,''
\newblock {\em ArXiv}, vol. abs/1802.01436, 2018.

\bibitem{article}
Tomasz Marciniak, Agata Chmielewska, Radoslaw Weychan, Marianna Parzych, and
  Adam Dabrowski,
\newblock ``Influence of low resolution of images on reliability of face
  detection and recognition,''
\newblock {\em Multimedia Tools and Applications}, vol. 74, 06 2013.

\bibitem{Li2019OnLF}
Pei Li, Loreto Prieto, Domingo Mery, and Patrick~J. Flynn,
\newblock ``On low-resolution face recognition in the wild: Comparisons and new
  techniques,''
\newblock {\em IEEE Transactions on Information Forensics and Security}, vol.
  14, pp. 2000--2012, 2019.

\bibitem{Chollet2017XceptionDL}
François Chollet,
\newblock ``Xception: Deep learning with depthwise separable convolutions,''
\newblock {\em 2017 IEEE Conference on Computer Vision and Pattern Recognition
  (CVPR)}, pp. 1800--1807, 2017.

\bibitem{Ford2019AdversarialEA}
Nic Ford, Justin Gilmer, Nicholas Carlini, and Ekin~Dogus Cubuk,
\newblock ``Adversarial examples are a natural consequence of test error in
  noise,''
\newblock in {\em ICML}, 2019.

\bibitem{Rusak2020IncreasingTR}
Evgenia Rusak, Lukas Schott, Roland~S. Zimmermann, Julian Bitterwolf, Oliver
  Bringmann, Matthias Bethge, and Wieland Brendel,
\newblock ``Increasing the robustness of dnns against image corruptions by
  playing the game of noise,''
\newblock {\em ArXiv}, vol. abs/2001.06057, 2020.

\end{thebibliography}

\end{document}